\DocumentMetadata{%
	pdfstandard=A-3u,
	pdfversion=1.7,
	lang=en-US,
}

\documentclass[nofoot,balance,colorlinks,upint,subscriptcorrection,varvw,mathalfa=cal=boondoxo,spanish,german,vietnamese,russian,greek]{asmeconf}

\hypersetup{%
	pdfauthor={John H. Lienhard},									  
	pdftitle={ASME Conference Paper LaTeX Template},                  
	pdfkeywords={ASME conference paper, LaTeX template, BibTeX style},
	pdfsubject = {Describes the asmeconf LaTeX template},			  
	pdflicenseurl={https://ctan.org/pkg/asmeconf},
}

\usepackage{graphicx} 
\usepackage{amsmath}
\usepackage{booktabs} 
\usepackage{comment}

\usepackage{array}
\usepackage{longtable}
\usepackage{rotating}
\usepackage{float} %
\usepackage{booktabs} %

\begin{document}




\title{Generative Machine Learning in Adaptive Control of Dynamic Manufacturing Processes: A Review
} 
 
%
%
%

\SetAuthors{%
	Suk Ki Lee\affil{},  
	Hyunwoong Ko\affil{}\CorrespondingAuthor{Hyunwoong.Ko@asu.edu}
	}

\SetAffiliation{}{School of Manufacturing Systems and Networks, Arizona State University, Mesa, AZ}


\maketitle





\begin{abstract}
Dynamic manufacturing processes exhibit complex characteristics defined by time-varying parameters, nonlinear behaviors, and uncertainties.
These characteristics require sophisticated in-situ monitoring techniques utilizing multimodal sensor data and adaptive control systems that can respond to real-time feedback while maintaining product quality.
Recently, generative machine learning (ML) has emerged as a powerful tool for modeling complex distributions and generating synthetic data while handling these manufacturing uncertainties.
However, adopting these generative technologies in dynamic manufacturing systems lacks a functional control-oriented perspective to translate their probabilistic understanding into actionable process controls while respecting manufacturing-specific constraints.
This review presents a functional classification of Prediction-Based, Direct Policy, Quality Inference, and Knowledge-Integrated approaches, offering an analytical perspective for understanding existing ML-enhanced control systems and incorporating generative ML.
The analysis of generative ML architectures within the established functional viewpoint demonstrates their unique control-relevant properties and potential to extend current ML-enhanced approaches where conventional methods prove insufficient.
This study then presents generative ML's potential for manufacturing control through decision-making applications, process guidance, simulation, and digital twins, while identifying critical research gaps: separation between generation and control functions, insufficient physical understanding of manufacturing phenomena, and challenges adapting models from other domains.
In response to these challenges and opportunities, the study proposes future research directions aimed at developing integrated frameworks that effectively combine generative ML and control technologies to address the dynamic complexities of modern manufacturing systems.
\end{abstract}

\section{INTRODUCTION}
Manufacturing processes are becoming increasingly complex and dynamic, driven by the demands for products with higher complexity, quality improvement, process efficiency, and manufacturing flexibility \cite{qu2019smart, tao2018data}. 
The integration of manufacturing and digital technologies in Industry 4.0 has further transformed traditional manufacturing paradigms \cite{lu2020digital,everton2016review,soori2024virtual}. 
These modern manufacturing environments face unprecedented challenges in maintaining consistent quality and optimal performance due to their inherent complexity, ranging from rapid parameter variations to unpredictable process dynamics and environmental uncertainties \cite{peres2020industrial,jan2023artificial}. 
Traditional control approaches, while providing a strong foundation for manufacturing automation, often struggle to fully address these challenges, particularly in highly dynamic and uncertain conditions.

The emergence of machine learning (ML) technologies has opened new possibilities for enhancing manufacturing control systems \cite{cai2023review}. 
Even though, in recent times, conventional ML approaches have already demonstrated significant improvements in process monitoring and control \cite{lough2020situ}, the recent advent of generative ML presents potentially transformative opportunities for manufacturing. 
Generative ML refers to the technologies that generate realistic data by understanding the original data's comprehensive distribution and underlying hidden patterns.
These generative ML technologies offer promising capabilities in handling complex, dynamic systems through their ability to learn, adapt, and generate scenarios in manufacturing environments.
However, adopting these generative ML capabilities in dynamic manufacturing systems lacks a functional control-oriented perspective that can efficiently translate their probabilistic understanding into actionable process controls while respecting manufacturing-specific constraints.
To address the challenge, we explore emerging pathways for integrating control methods and generative ML-enhanced adaptive systems in modern manufacturing environments. 

To structure this exploration, we adopt a systematic approach that progressively builds from the necessary foundation to future direction, as follows.
This review provides a comprehensive overview of dynamic manufacturing processes, including their complex characteristics, in-situ monitoring approaches, and control requirements in Section 2. 
We then examine the current landscape of ML-enhanced adaptive control in manufacturing, introducing a functional classification to analyze various methodologies and their industrial applications in Section 3. 
Next, we introduce key generative ML technologies and their control-relevant properties in Section 4. 
Further, we analyze the current integration status of generative ML in adaptive control, analyzing applications in decision-making, process guidance, simulation, digital twins, transferable approaches from related domains, and identifying critical research gaps in Section 5. 
Finally, we conclude with a discussion of future research directions for integrating generative ML with manufacturing control systems in Section 6.


\section{DYNAMIC MANUFACTURING PROCESSES}
Modern manufacturing involves complexities and uncertainties stemming from its inherently dynamic nature, distinguishing it from traditional manufacturing systems.
Addressing these challenges requires the adoption of sophisticated real-time monitoring and advanced control strategies, capable of responding effectively to rapidly evolving conditions and escalating process complexities. 
Such strategies are essential to achieve consistently high-quality production and process stability.
This section explores key characteristics inherent to dynamic manufacturing processes, emphasizing the critical roles of in-situ monitoring systems and the essential control aspects.

\subsection{Complex Process Characteristics}
Rapid and continuous parametric variations, nonlinear behaviors, and inherent uncertainties characterize dynamic manufacturing processes. 
These factors can significantly affect product quality and process efficiency.
For example, variations of process parameters in the laser-based additive manufacturing (AM) processes, such as laser power, scanning speed, melting location, and material properties, vary continuously. 
These variations lead to complex interactions challenging control systems \cite{everton2016review,cai2023review}.
Often, these variations result in nonlinear behaviors, where minor changes in input parameters can disproportionately affect melt pool geometry and material properties, as evidenced by thermographic and high-speed imaging studies \cite{zhang2018extraction}.
Process uncertainties, such as inconsistencies in material properties or environmental factors, further complicate manufacturing control. 
In laser-based processes, phenomena such as plume and spatter formation introduce unpredictable melt pool behavior, emphasizing the necessity of robust monitoring systems \cite{zhang2018extraction}.

Emerging manufacturing practices, including multi-robot systems, manufacturing in extreme environments such as in-space manufacturing, and advanced semiconductor manufacturing processes, introduce unprecedented challenges that substantially intensify these systems' dynamic nature and control requirements beyond modern manufacturing processes' inherent complexities.
The multi-robot systems escalate process complexity through requirements for precise synchronization, dynamic path planning, and coordinated behavior control \cite{zhang2022aerial}.
Furthermore, hybrid robotic systems, such as collaborative aerial-ground multi-robot systems, intensify these complexities through disparate motion capabilities, diverse sensing modalities, and asynchronous operations needing harmonization \cite{krizmancic2020cooperative}.
Manufacturing in space introduces unique complexities due to microgravity conditions that fundamentally alter material flow behaviors, thermal gradients, and solidification mechanisms \cite{sacco2019additive}.
Additionally, extreme space conditions, including radiation exposure, vacuum, and temperature fluctuations, significantly impact material properties and process stability \cite{zocca2022challenges}.

Addressing these challenges demands adaptive control strategies integrated with sophisticated in-situ monitoring techniques to facilitate real-time adjustments and maintain process stability and product quality \cite{cai2023review}.

\subsection{In-Situ Monitoring in Manufacturing}
In-situ monitoring has become integral to dynamic manufacturing processes. 
Leveraging ML with in-situ data reveals data-driven, previously unseen information regarding phenomena characterized by complexities and uncertainties inherent in manufacturing processes \cite{ko2022spatial, lee2024amtransformer}. 
Continuous monitoring facilitates the early detection of anomalies and process drifts, thereby preventing defects and promoting consistent component quality.

Advanced manufacturing systems utilize diverse monitoring technologies. 
For example, optical imaging and thermography reveal surface characteristics and temperature distributions, capturing critical fluctuations indicative of energy absorption and cooling rate variations. 
High-speed optical and thermal imaging, in particular, have proven effective in capturing the dimension and fluctuations of melted regions, thermal profiles, material deposition rates, and spatter formation at the sites of energy-material interact \cite{zhang2018extraction,everton2016review,cai2023review,mitchell2020linking,lane2020process,lee2024amtransformer}.
These measurements serve as key indicators of process health. 
Optical emission spectroscopy analyzes spectral emissions, identifying porosities, material composition changes, and plasma characteristics \cite{lough2020situ,montazeri2020process}. 
Acoustic monitoring detects characteristic process sounds and potential defects \cite{tempelman2022detection,kononenko2023situ}.

Advancements in in-situ monitoring in manufacturing have shifted toward multimodal data integration approaches.
Investigating correlations between different modality data showed a possibility that multimodal fusion could effectively combine the strengths of different monitoring systems \cite{yang2022investigating}.
The multimodal sensor framework proposed by Chen has been shown to improve defect identification and quality assurance by capturing complementary information from various sensing data incorporating acoustic, thermal, displacement, and visual data \cite{chen2024multi}.  
Advanced multimodal integration through unsupervised contrastive learning that compresses high-dimensional sensor data into low-dimensional representational spaces, creating a flexible framework adaptable to diverse manufacturing environments \cite{mckinney2025unsupervised}.
These advancements enable adaptive, data-driven control in complex manufacturing processes, though successful integration of in-situ monitoring with control systems remains essential for maintaining high product quality through real-time feedback loops \cite{tao2018data}.

\subsection{Control Requirements for Dynamic Processes}
Abundant in-situ monitoring data yield high-resolution, temporally and spatially resolved information about the evolving dynamics of manufacturing processes. 
This information can be systematically translated into desired manufactured outcomes through the implementation of appropriate control actions within the dynamic processes environment.
To enable this transformation, dynamic manufacturing systems must be designed to respond adaptively to real-time process variations rather than operate based on fixed rules or predetermined routines.
Enhanced interpretation of heterogeneous in-situ data, coupled with real-time or near-real-time adjustments to process parameters, is essential for improving product quality and process stability, which are core objectives of advanced manufacturing systems.  
Achieving these objectives necessitates control frameworks capable of dynamically adapting to changing conditions while managing the uncertainties captured through in-situ monitoring.

Control systems for dynamic manufacturing processes require three essential capabilities to effectively respond to complexities revealed through in-situ monitoring.
First, adaptation mechanisms must swiftly adjust process parameters based on feedback to maintain process stability during rapidly changing conditions. 
\cite{chen2020data,meng2020machine}.
Second, control systems must handle inherent uncertainties, including sensor noise, material variability, and environmental fluctuations in manufacturing, ensuring sustained reliability despite potential performance degradation from noise or sensor drift \cite{jourdan2021reliability,wang2018situ,tao2018data}.
Third, quality maintenance requires translating process signatures into corrective actions, predictive interventions before defects materialize, and continuous feedback loops that iteratively optimize quality-critical parameters. 
\cite{everton2016review, lu2020digital}.
These capabilities are achievable through ML-enhanced adaptive control approaches, incorporating advanced data analytics and adaptive algorithms, effectively mitigating process variations consistently \cite{qu2019smart}.

\section{ML-Enhanced Adaptive Control in Manufacturing}
Traditional adaptive control methods have provided foundational approaches to managing manufacturing processes, but their reliance on explicit rules and mathematical models limits their effectiveness with complex, nonlinear, and high-dimensional systems typical in modern manufacturing \cite{hespanha2003overcoming, fang2022process}.
Applying ML to adaptive control offers enhancement by leveraging data-driven approaches that can learn complex patterns and adapt to changing conditions without extensive domain modeling. 
This section explores how ML enhances adaptive control in manufacturing, first providing an overview of its advantages and then exploring specific control paradigms in manufacturing.

\subsection{Overview of ML-Enhanced Adaptive Control}
The integration of ML into adaptive control, so-called ML-enhanced adaptive control, utilizes data-driven ML methodologies to improve manufacturing process control. The integration offers unprecedented capabilities in dynamic manufacturing environments, extending beyond knowledge-driven rule-based approaches.
ML provides several key advantages for adaptive control in manufacturing settings.

First, ML-enhanced adaptive control systems excel in discriminative modeling and prediction.
ML techniques, such as neural networks (NNs), are highly effective in learning direct mappings between process states and optimal control actions. This capability facilitates precise parameter adjustments tailored to current process conditions without requiring complete modeling of the entire data generation process \cite{inyang2022learn, kuang2020precise, moradimaryamnegari2025neural}. 
Second, ML's strength in feature-based learning significantly enhances control effectiveness.
ML methods efficiently extract and prioritize relevant features from complex sensor data, enabling targeted and effective control decisions without necessitating comprehensive modeling of all process variables \cite{humfeld2021machine}.
Third, ML enhances the efficiency of adaptive control systems, frequently excelling at identifying optimal control solutions.  
In contrast to simulation-based adaptive approaches that typically require extensive modeling and validation efforts, ML methods achieve robust control performance with relatively modest training data \cite{liu2024deep,chen2025real}. 
Consequently, ML-based adaptive controls are particularly powerful for manufacturing environments with limited process knowledge or capabilities or to perform high-fidelity simulations.

ML-integrated adaptive control has been adopted across different manufacturing fields. 
Sectors such as AM and semiconductor manufacturing have been early adopters, leveraging ML to address complex control challenges unique to their processes. \cite{wang2018deep, lee2018industrial}.
The adoption strategy increasingly combines real-time control with virtual modeling approaches, where ML-enhanced controllers operate in conjunction with digital twin (DT) frameworks. 
Recent advances have focused on transitioning from offline to online applications, enabling systems to adapt in real time while maintaining stability guarantees \cite{zheng2023adaptive,he2024self}.
Simultaneously, integration with DT technologies has emerged as a powerful approach for comprehensive process optimization \cite{chiurco2023data}. 

\subsection{ML-Enhanced Control Methodologies and Applications}
ML-enhanced adaptive control in manufacturing can be categorized by examining how information is processed to generate control decisions rather than by specific algorithms employed.
This study presents a functional classification with four distinct approaches: (1) Prediction-Based control, where ML forecasts future system states to optimize decisions; (2) Direct Policy control, where ML learns to map system states directly to control actions; (3) Quality Inference control, where ML estimates unmeasurable quality parameters to guide process adjustments; and (4) Knowledge-Integrated control, where ML combines data-driven learning with physics-based constraints.
This categorization offers a point of view for understanding how different ML integration methods address manufacturing challenges based on their control objectives and information processing paradigms.

Prediction-Based control approaches utilize explicit forecasting of system dynamics to anticipate future states and optimize control decisions \cite{shen2019learning,zhang2019process}. 
Such methods enhance model predictive control (MPC) by integrating ML models trained on manufacturing process data. 
While maintaining the MPC structure, they leverage ML techniques to create more accurate predictive models.
These approaches enable precise process control by anticipating potential deviations before they develop into defects. 
Research across AM technologies demonstrates the versatility of this approach: Shen et al. applied 3D CNN-autoencoder architectures to predict and compensate for geometrical deformations in polymer printing \cite{shen2019learning}; and Zhang et al. utilized CNNs to analyze melt-pool images and predict porosity formation in direct energy deposition processes \cite{zhang2019process}. 
These applications collectively highlight how Prediction-Based control enables manufacturers to move from reactive correction to proactive intervention across diverse AM technologies and materials.

Direct Policy control approaches learn mappings directly from system states to control actions without explicit process modeling, bypassing other approaches' prediction and optimization steps \cite{mattera2024optimal, kuhnle2021designing,boydon2023deep}.
Reinforcement learning (RL) is the primary ML, with algorithms learning optimal control strategies through reward-based feedback. 
Mattera et al. demonstrated RL techniques that optimize parameters like material feed rate and heat input in real time for improved dimensional accuracy 
\cite{mattera2024optimal}.
In computer numerical control (CNC) machining, Kuhnle et al. developed multi-agent RL systems that simultaneously manage parameter control and production scheduling to reduce energy consumption while maintaining throughput. 
\cite{kuhnle2021designing}.
In semiconductor manufacturing, Boydon et al. implemented deep learning agents trained on Markov decision process solutions for dynamic production control, generating near-optimal policies in a fraction of the computational time while significantly improving cycle times compared to traditional dispatching rules \cite{boydon2023deep}.
These applications highlight Direct Policy control's effectiveness for manufacturing processes with complex dynamics where learning from experience proves more practical than developing explicit models, enabling adaptation to changing conditions and capturing nonlinear relationships conventional approaches cannot represent.

Quality Inference control uses ML to estimate unmeasurable process parameters and quality characteristics in real-time, focusing on current system states rather than future predictions \cite{kang2009virtual,tin2022virtual}.
While Prediction-Based control forecasts future system behavior, this approach develops ML-based inference models that function as virtual measurement instruments, transforming readily available process signals into accurate estimates of critical quality metrics that would typically require specialized physical measurement equipment. 
Especially in semiconductor manufacturing, where direct measurements are often prohibitively expensive, time-consuming, or physically impossible during the process, Quality Inference control has gained significant traction. 
Researchers such as Kang et al. and Tin et al. have demonstrated various ML techniques, from support vector regression (SVR) and NNs for wafer thickness estimation during chemical mechanical planarization to CNN-based systems for predicting photolithography overlay errors, achieving sub-nanometer accuracy and enabling immediate process optimization
\cite{kang2009virtual,tin2022virtual}.
These studies demonstrate how Quality Inference control enables manufacturers to monitor and maintain product quality through real-time process adjustments while significantly reducing the operational and financial burden of actual metrology.

Knowledge-Integrated control embeds physical laws or domain expertise directly into ML model architectures, fundamentally guiding how ML processes information \cite{zheng2023physics,liao2023hybrid}. 
This approach enforces physical constraints within the learning process itself. 
Zheng and Wu demonstrated this with physics-informed recurrent networks for nonlinear systems, integrating physical laws with online parameter estimation \cite{zheng2023physics}.
Liao et al. applied similar principles to AM, combining thermal imaging data with physical laws to predict temperature distributions and identify unknown parameters \cite{liao2023hybrid}.
These approaches bridge the gap between established physical models and data-driven learning, guiding complex processes where neither purely theoretical nor purely empirical methods prove alone. 
While showing promising results, the application of such techniques in comprehensive manufacturing control systems remains an emerging field with significant development potential.


Table \ref{table:ml_enhanced_control} summarizes these ML-enhanced approaches across various manufacturing processes based on this study's classification.
This classification shows how ML integration enables a shift from reactive to predictive control approaches, offering a structured framework for understanding how different methods address manufacturing challenges with enhanced adaptability in dynamic environments.
Although the integration of ML into adaptive control systems has addressed many traditional challenges, significant limitations persist across all approaches. 
Because of conventional ML models' over-reliance on training data distributions, they have limited ability to identify hidden patterns and uncertainty. 
This restriction hampers their capacity to uncover latent dynamics essential for anomaly prediction and complex process optimization.
While the integration of physical laws or domain expertise reduces epistemic uncertainty, they cannot fully address all uncertainties in manufacturing processes, and aleatoric uncertainty, such as material heterogeneity and measurement uncertainty, remains challenging \cite{mahadevan2022uncertainty}. 
These probabilistic characteristics require more sophisticated modeling approaches. 
Generative ML technologies, discussed next, offer solutions through their inherent probabilistic frameworks.

\begin{table*}[tbp]
\caption{SUMMARY OF ML-ENHANCED CONTROL APPROACHES IN MANUFACTURING}\label{table:ml_enhanced_control}
\centering
\small
\renewcommand{\arraystretch}{1}
\begin{tabular*}{\textwidth}{@{\extracolsep{\fill}}p{2.8cm}p{2.8cm}p{3.3cm}p{4.8cm}p{2.5cm}}
\toprule
\textbf{ML-Enhanced \newline Control Approach} & \textbf{ML Technique} & \textbf{Manufacturing \newline Application} & \textbf{Key Features} & \textbf{Reference} \\
\midrule
%
%
\textbf{Prediction-Based Control} & 3D CNN with \newline Autoencoder & Polymer AM (FDM) & Real-time error detection; Quality optimization & Shen (2019)~\cite{shen2019learning} \\

& CNN & Direct Energy Deposition & Porosity monitoring; Process parameter control & Zhang (2019)~\cite{zhang2019process} \\
\midrule
\textbf{Direct Policy \newline Control} & Deep RL & Wire Arc AM & Reduced-order modeling; Sim-to-real transfer & Mattera (2024)~\cite{mattera2024optimal} \\

& Deep RL & CNC Machining & Tool wear prediction; Feed rate adjustment & Kuhnle (2021)~\cite{kuhnle2021designing} \\

& Deep Learning Agents & Semiconductor & Near-optimal policy generation & Boydon (2023)~\cite{boydon2023deep} \\
\midrule
\textbf{Quality Inference Control} & SVR + NN & Semiconductor & Real-time process monitoring; Parameter optimization & Kang (2009)~\cite{kang2009virtual} \\

& CNN & Semiconductor & Overlay error prediction & Tin (2022)~\cite{tin2022virtual} \\
\midrule
\textbf{Knowledge-Integrated Control} & Physics-informed RNN & Chemical Process Control & Physics-embedded learning; Online parameter estimation & Zheng (2023)~\cite{zheng2023physics} \\

& Physics-informed NN & Direct Energy Deposition & Thermal field prediction; Unknown parameter identification & Liao (2023)~\cite{liao2023hybrid} \\
\bottomrule
\end{tabular*}
\end{table*}


\section{GENERATIVE ML TECHNOLOGIES: CONTROL-RELEVANT PROPERTIES} \label{Section_GenerativeML}
Generative ML has emerged as a powerful tool for solving complex problems across many domains \cite{goodfellow2016deep}. These technologies aim to understand and model the underlying data distributions from observations, enabling the generation of new samples or predictions that reflect the learned patterns \cite{bishop2006pattern}. Their generative capabilities open numerous possibilities for improving control in manufacturing processes.
This section reviews key generative ML technologies and how those methods can be utilized for process control, especially in manufacturing processes.
We focus on four major architectures that have shown promising results in control applications: Variational Autoencoders (VAEs), Generative Adversarial Networks (GANs), Transformer-based models, and Diffusion models.
These architectures represent the evolution of generative ML, each bringing unique strengths to process control challenges.
Through this section, we examine each architecture's core mechanisms and their control-relevant properties to illuminate how these technologies can be effectively integrated into control systems.

\subsection{Variational Autoencoders (VAEs)}
VAEs are one type of technology that opens the initial advancement of generative models by leveraging stochastic concepts for inference and combining them with deep NNs \cite{kingma2013auto}.
VAEs employ the mean and variance when mapping the input data into encoded latent representation and reconstructing it through a decoding process.
The encoder models input data as a probability distribution in latent space, and employs the reparameterization trick, as shown in Equations \ref{eq:vae_encoder} and \ref{eq:vae_reparam}.
\begin{equation}
q_\phi(z|x) = \mathcal{N}(z; \mu_\phi(x), \sigma_\phi^2(x) I)
\label{eq:vae_encoder}
\end{equation}
\begin{equation}
z = \mu_\phi(x) + \sigma_\phi(x) \cdot \epsilon, \quad \epsilon \sim \mathcal{N}(0, I)
\label{eq:vae_reparam}
\end{equation}
\noindent, where $x$ is the input data, $z$ is the latent variable, $\mu_\phi(x)$ and $\sigma_\phi^2(x)$ are the mean and variance outputs from the encoder network, and $\epsilon$ is random noise sampled from a standard normal distribution.
This approach enables optimization of the evidence lower bound, as shown in Equation \ref{eq:vae_elbo}.
\begin{equation}
\mathcal{L}(\theta, \phi) = \mathbb{E}_{q_\phi(z|x)} [\log p_\theta(x|z)] \\
- D_{\text{KL}}(q_\phi(z|x) || p(z))
\label{eq:vae_elbo}
\end{equation}
\noindent, where $\theta$ and $\phi$ are the decoder and encoder parameters, $\mathbb{E}$ denotes expectation, $p_\theta(x|z)$ is the decoder likelihood, and $q_\phi(z|x)$ is the encoder distribution. The first term maximizes reconstruction quality, while the second term with prior $p(z)$ and Kullback-Leibler divergence $D_{KL}$ serves as regularization.
This probabilistic foundation enables VAEs to capture uncertainty inherently in their architecture, differentiating them from deterministic autoencoders \cite{doersch2016tutorial}.

VAEs offer several key properties that make them particularly valuable in control applications: (1) Their latent space representation provides a powerful framework for process control systems.
This latent space captures physically meaningful relationships of process dynamics, allowing a reduced-dimensional space that enables efficient control and monitoring \cite{watter2015embed}. 
The reduction maintains critical correlations between process variables while eliminating redundant information, leading to more tractable optimization problems in model predictive control frameworks. 
Notably, the learned representations often align with physically meaningful process parameters, enhancing interpretability.
(2) The second key property stems from VAE's probabilistic framework, which inherently quantifies the uncertainty of the input state. 
This capability is crucial for robust control, as the encoded uncertainty information helps identify regions where the model might have high possibilities of anomalous states, enabling more cautious control actions in these areas. 
Additionally, it provides confidence bounds on model outputs for more reliable control decisions.
(3) VAEs have the capability to generate realistic process scenarios containing correlations between process variables. 
This property enables exploring the possible range of system behaviors and provides opportunities for testing control strategies and process parameter optimizations, fostering better understanding and management of system behaviors \cite{hewing2019cautious}.

\subsection{Generative Adversarial Networks (GANs)}
GANs are generative modeling approaches designed to generate new samples that closely resemble real data \cite{goodfellow2014generative}. 
GANs consist of two NNs, a generator and a discriminator, trained in a competitive framework.
The generator learns to produce synthetic data that mimics the training distribution, while the discriminator attempts to distinguish between real and generated samples. 
During training, the generator aims to improve its output to fool the discriminator, while the discriminator tries to classify real data versus fake data correctly.
This adversarial process is formalized in Equation \ref{Eq:gan_objective}.
\begin{multline}
\min_G \max_D V(D, G) = \mathbb{E}_{\mathbf{x} \sim p_{\text{data}}(\mathbf{x})}\left[\log D(\mathbf{x})\right] \\
+ \mathbb{E}_{\mathbf{z} \sim p_{\mathbf{z}}(\mathbf{z})}\left[\log(1 - D(G(\mathbf{z})))\right]
\label{Eq:gan_objective}
\end{multline}
\noindent, where $G$ represents the generator, $D$ is the discriminator , $p_{\text{data}}(\mathbf{x})$ is the real data distribution, and $p_{\mathbf{z}}(\mathbf{z})$ is the prior noise distribution, typically Gaussian.
This adversarial process results in a generator capable of producing high-quality, realistic data.

In process control contexts, GANs offer three distinctive properties:
(1) Their implicit distribution learning capability differs from VAEs, which require explicit probability calculations, as GANs learn to generate realistic data directly.
This implicit approach is especially effective for modeling complex, high-dimensional process data where explicit probabilistic modeling is too complicated and infeasible.
GANs can produce synthetic data that adheres to physical constraints without requiring these constraints to be explicitly encoded.
(2) GANs can augment datasets by generating synthetic process data while maintaining complex statistical relationships found in the original data \cite{wang2025mcgan}.
This capability addresses the common challenge of limited operational data in real-world industrial circumstances.
The generated samples aim to capture statistical relationships between process variables, which can support control system development.
These synthetic data can be used to improve the robustness and accuracy of ML-based control systems without requiring costly physical trials.
(3) GANs' adversarial mechanism provides unique advantages for anomaly generation \cite{sabuhi2021applications, kumbhar2023deepinspect}. 
The discriminator acts as an adaptive loss function, evolving to find subtle process anomalies or unrealistic behaviors continuously. 
This competitive optimization enhances the quality of generated data and enables GANs to simulate rare fault cases. 
These fault scenarios allow control systems to be rigorously tested and trained to handle unexpected events, which could improve their reliability and robustness in real-world applications.

\subsection{Transformer-based Models}
While Transformers are not inherently generative, Transformers' exceptional sequential modeling capabilities have enabled the development of powerful generative models such as GPT and BART \cite{radford2018improving, lewis2019bart}.
The Transformer architecture brings about a paradigm shift in sequential modeling approaches through its innovative self-attention mechanism, which offers several key properties beneficial for control systems \cite{vaswani2017attention}.
This attention mechanism lies at the core of the architecture, enabling the direct modeling of dependencies regardless of their sequential distance, as shown in Equation \ref{Eq:Attention}.
\begin{equation}\label{Eq:Attention}
    \text{Attention}(Q, K, V) = \text{softmax}\left(\frac{QK^T}{\sqrt{d_k}}\right)V
\end{equation}
\noindent, where $Q$ represents queries as the current process state, $K$ represents keys as reference points in data history, $V$ represents values as process parameters or data associated with reference points, and $d_k$ is the dimension of the keys.
The resulting attention output is a weighted sum of values, where weights reflect the relevance of each reference point to the current process state.
Unlike traditional recurrent architectures, Transformers process entire sequences in parallel, allowing them to compute relationships between all elements simultaneously. 
To retain temporal information, Transformers incorporate positional encodings. At the same time, the multi-head attention mechanism enhances the model’s ability to capture diverse relationships by attending to different aspects of the input data simultaneously. 
This allows the Transformer to model intricate dependencies within the sequence more effectively than single-head attention approaches.

Transformers are particularly advantageous for control systems, as they help in understanding complex dependencies due to their unique properties: 
(1) Transformers have the ability to reveal and understand global dependencies across sequential inputs by utilizing attention mechanisms, representing a significant advantage in process control \cite{lee2024amtransformer}.
This property enables the identification of relationships across different times, while traditional sequential models struggle with long-range dependencies. 
Transformers could capture correlations between events regardless of their temporal separation, making them particularly valuable in processes that require comprehensive temporal analysis. 
Based on this property, transformer models could enable more accurate predictions and anticipatory control strategies, enhancing system performance.
(2) Multi-head attention enables a deeper understanding of input data by allowing the model to attend to various aspects of the sequence simultaneously  \cite{lee2024amtransformer,chen2021decision}. 
Each attention head focuses on a unique subset of relationships within the input, capturing both fine-grained and broad patterns in the data. 
This capability improves the model's ability to represent complex process dynamics and enables a more comprehensive system understanding, essential for control systems requiring high adaptability and precision.
(3) Transformers’ attention mechanisms can provide interpretability, offering insights into how the model weighs different temporal relationships when making predictions or control decisions \cite{chen2021decision}. 
The attention weights reveal which prior input elements most strongly influence the current output, enhancing the transparency of the model's decisions. 
This interpretability is not only crucial for validating control decisions but also aids in identifying critical process relationships that may not be apparent through traditional methods. 
Furthermore, the attention patterns can uncover unexpected dependencies in the process, supporting both system understanding and the refinement of control strategies.

\subsection{Diffusion Models}
Diffusion models are generative models that are based on the principle of gradually denoising data through a learned reverse diffusion process \cite{ho2020denoising,nichol2021improved}. 
The diffusion framework consists of two key steps: the forward process and the reverse process.
In the forward process, the model gradually adds Gaussian noise to the original data across the steps, which means diffusing data, as shown in Equation \ref{Eq:diff_forward}.
\begin{equation}
q(\mathbf{x}_t | \mathbf{x}_{t-1}) = \mathcal{N}(\mathbf{x}_t; \sqrt{1 - \beta_t} \mathbf{x}_{t-1}, \beta_t \mathbf{I}), \quad t \in [1,T]
\label{Eq:diff_forward}
\end{equation}
\noindent, where $\mathbf{x}_0$ represents the original manufacturing data in manufacturing context, $\mathbf{x}_t$ is the noised data at timestep $t$, and $\beta_t$ is the noise schedule controlling the diffusion rate.
Then, the model learns the data distribution through a reverse process that reconstructs the original data using iterative denoising step by step, as defined in Equations \ref{Eq:diff_reverse} and \ref{Eq:diff_learning}.
\begin{equation}
p_\theta(\mathbf{x}_{t-1} | \mathbf{x}_t) = \mathcal{N}(\mathbf{x}_{t-1}; \boldsymbol{\mu}_\theta(\mathbf{x}_t, t), \boldsymbol{\Sigma}_\theta(\mathbf{x}_t, t))
\label{Eq:diff_reverse}
\end{equation}
\begin{equation}
\mathcal{L}_{\text{simple}} = \mathbb{E}_{\mathbf{x}_0, \boldsymbol{\epsilon}, t} \left[ \|\boldsymbol{\epsilon} - \boldsymbol{\epsilon}_\theta(\mathbf{x}_t, t) \|^2 \right]
\label{Eq:diff_learning}
\end{equation}
\noindent, where $p_\theta$ represents the learned reverse process with parameters $\theta$, $\boldsymbol{\mu}_\theta$ and $\boldsymbol{\Sigma}_\theta$ are the predicted mean and covariance, and $\boldsymbol{\epsilon}_\theta$ is a NN that predicts the noise component added during the forward process.
This iterative refinement allows diffusion models to generate results that are not only high in fidelity but also adaptable to complex process constraints, making them particularly effective for tasks requiring gradual state transitions.
This approach is fundamentally different from both VAEs and GANs, as it directly models the gradient of the data distribution through score matching \cite{song2020score}.

The unique iterative nature of diffusion models offers several advantageous properties from the process control perspective: 
(1) The iterative denoising process of diffusion models enables process optimization by generating highly realistic process trajectories \cite{janner2022planning, giannone2023aligning}.
The gradual refinement approach through multiple steps allows for more precise control over the generation process, which differs from single-step generation approaches in other models.
This capability is especially valuable in planning smooth transitions between process states while adhering to physical constraints, making them ideal for processes requiring controlled state evolution.
(2) Diffusion models provide robust probabilistic state estimation through their inherent uncertainty quantification capabilities, especially through their probabilistic formulation of the denoising reverse process \cite{tang2024adadiff}. 
Each step of the reverse process provides probabilistic estimates, allowing for comprehensive uncertainty quantification throughout the generation process.
(3) Simultaneously, the denoising steps can be guided by incorporating control objectives or physical constraints, ensuring that the probabilistic estimates remain consistent with the desired outcomes \cite{dhariwal2021diffusion,song2020score, bar2023multidiffusion, giannone2023aligning}. 
This property enables risk-aware control strategies that can account for varying levels of uncertainty at different stages of process evolution. 
Additionally, the models can generate multiple plausible trajectories from the same initial conditions, supporting robust control design through scenario analysis and improving reliability and adaptability in dynamic systems.

\begin{figure}[htbp]
\vspace*{4pt}
\centerline{\includegraphics[width=\columnwidth]{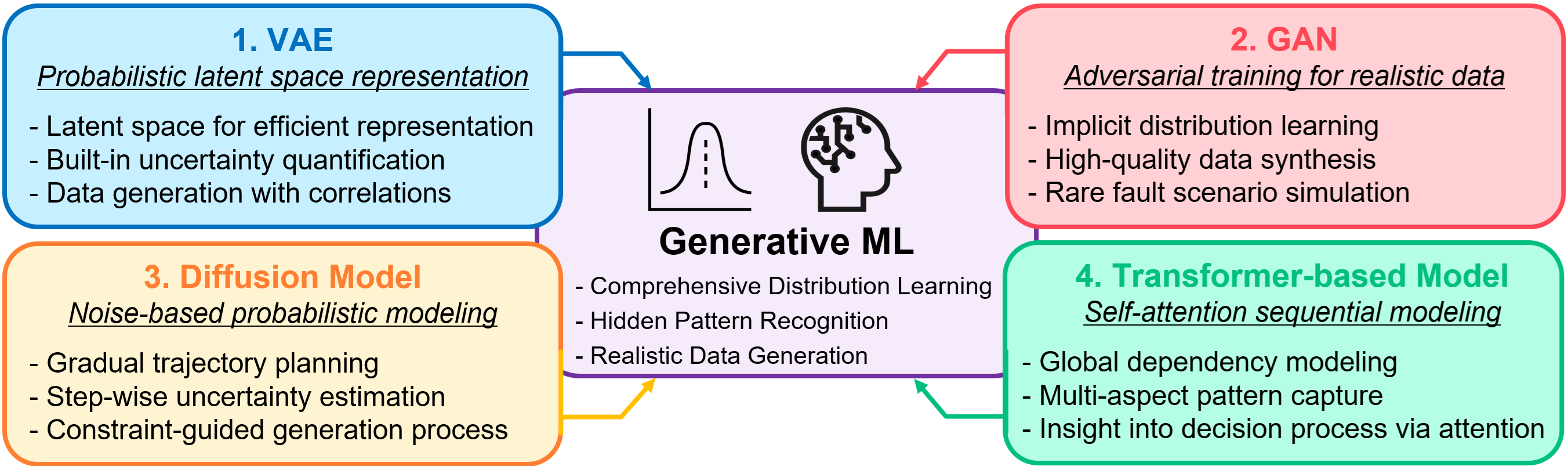}} 
\caption{Generative ML Overview with Model-specific properties.}
\label{fig_GenAI_Overview_with_model_properties} 
\end{figure}

Figure \ref{fig_GenAI_Overview_with_model_properties} shows the common characteristics of generative ML models and their distinctive properties for process control.
These generative techniques offer unique advantages in learning comprehensive data distributions, enabling them to uniquely capture hidden patterns and relationships within process data.
By mitigating input biases, generative models produce outputs that are not only realistic but also capture subtle, latent structures often overlooked by traditional methods.
This ability makes generative ML highly effective in developing robust and adaptable solutions, particularly in data-intensive and complex control scenarios.

\section{CURRENT INTEGRATION STATUS OF GENERATIVE ML IN ADAPTIVE CONTROL}
Integrating generative ML's capabilities reviewed in Section \ref{Section_GenerativeML}  with manufacturing control creates synergy for adaptive control systems.
Nevertheless, leveraging generative ML in manufacturing processes remains an emerging research area.
This section examines the current research landscape of generative ML in adaptive manufacturing control, categorizing contributions into generative ML applications for manufacturing decision-making and parameter optimization, data-driven manufacturing simulation and digital twin (DT) construction, and transferable approaches from related domains. 
Furthermore, it points out significant research gaps, underscoring challenges and opportunities for future investigation in this area.

\subsection{Generative ML for Manufacturing Decision-Making and Process Guidance}
Current research of generative ML-integrated approaches in manufacturing primarily aligns with Prediction-Based and Direct Policy control approaches reviewed in Section 3.
Li et al. demonstrate Prediction-Based control through their GAN-Gated Recurrent Unit (GRU) architecture for welding systems. 
Their model generates future weld pool images based on torch speed adjustments, creating a human-centered MPC system where operators visualize consequences before implementation \cite{li2025generative}.
This conditional GAN framework, enhanced with GRUs for temporal modeling, captures relationships between speed variations and weld pool morphology while preserving human judgment in the loop. 
The system effectively reduces operator skill requirements by transforming adaptive control through future-state forecasting in decision processes.
In scheduling optimization, a transformer architecture with deep RL by Li et al. exemplifies Direct Policy control \cite{li2024transformer}.
Their system maps production states directly to scheduling decisions in dynamic environments, with the transformer capturing temporal dependencies across manufacturing workflows.
%
%
While focused on scheduling tasks rather than direct process control, this study demonstrates the possibility of a generative ML model, a Transformer, that can be effectively integrated into manufacturing control systems.
These applications demonstrate how generative ML enhances manufacturing decision-making through both prediction and policy learning.

\subsection{Generative ML for Manufacturing Simulation and Digital Twins}
Generative ML demonstrates significant potential for simulation and DTs in manufacturing. 
Existing studies primarily align with the Quality Inference control approaches, with potential extensions toward Knowledge-Integrated approaches.
The study conducted by Mu et al. explores how generative ML models can be effectively combined to simulate complex manufacturing processes \cite{mu2024online}.
In this research, a diffusion-based generative ML framework integrates a Vector Quantized VAE coupled with GANs for spatial feature extraction, while a Recurrent NN (RNN) handles the fusion of time-scale results. 
This hybrid approach generates spatially accurate distortion field prediction in wire arc additive manufacturing, otherwise difficult to measure during manufacturing, enabling anticipation of anomalies before they materialize.
Kim et al. employed a conditional GAN to predict surface morphology in directed energy deposition based on process parameters \cite{kim2023virtual}.
Their model efficiently generated realistic surface texture predictions without requiring computationally intensive physical simulations, providing a virtual simulation for quality prediction in AM.
Such capabilities are fundamental to DT construction, where digital representations must accurately reflect physical processes to provide meaningful insights for manufacturing control.
These studies demonstrate how generative models can effectively infer characteristics that are crucial information for control decisions.
While current generative approaches remain largely data-driven, incorporating physics-based constraints, as demonstrated in Knowledge-integrated control approaches, would yield more physically consistent predictions.
These physics-informed generative models could produce more reliable data for decision-making, enabling virtual validation of control strategies before physical deployment and accelerating the development of adaptive manufacturing systems.

\subsection{Transferable Approaches from Related Domains}
Generative ML applications in other domains could offer insights for adaptive control and decision-making while not directly related to manufacturing. 
Robotics, in particular, addresses similar challenges, including real-time trajectory planning, multi-task adaptability, and feedback-driven adjustments \cite{chi2023diffusion,janner2022planning,brohan2022rt,shridhar2023perceiver,chen2023transformer,meo2021multimodal,kapelyukh2023dall}. 
Various generative architectures deployed in robotics demonstrate transferable solutions for manufacturing systems.
Additionally, while not traditionally classified as generative ML, large language models (LLMs) have demonstrated capability in handling complex planning and decision-making for complex systems \cite{xia2023towards}.
Approaches from related domains can be bridged to manufacturing control by extracting transferable insights from generative ML applications in these fields.

Diffusion models in robotics demonstrate how their core capabilities can transfer to manufacturing tasks like dynamic parameter adjustment and process adaptation. 
Chi et al. apply diffusion models for real-time visuomotor policy learning in robotics, which can inform tasks like laser path optimization in manufacturing \cite{chi2023diffusion}. 
For long-horizon planning challenges, Janner et al. extended diffusion models to create scalable frameworks adaptable to manufacturing job scheduling and coordination \cite{janner2022planning}. 
Kapelyukh et al. demonstrate object placement in unstructured environments, highlighting the potential for modular and flexible workflows \cite{kapelyukh2023dall}. 

In robotics, transformers showcase flexibility in handling different tasks due to their capability to process complex, high-dimensional data and multimodal inputs. 
Brohan et al. demonstrate end-to-end task handling in robotics, offering insights for adaptive workflows like assembly and inspection in manufacturing \cite{brohan2022rt}. 
Decentralized multi-robot path planning approach of Chen et al. aligning with distributed manufacturing needs \cite{chen2023transformer}. 
Emphasize multimodal integration from Shridhar et al.'s work, suggesting applications in complex manufacturing tools \cite{shridhar2023perceiver}. 
These studies highlight the transformative potential of transformers for multi-task adaptability in manufacturing.

LLMs demonstrate strong potential for dynamic task allocation and workflow adaptation in distributed environments. 
In the study of Xia et al., LLMs are used for real-time task planning and intelligent resource allocation in modular production systems \cite{xia2023towards}. 
Distributed manufacturing systems, where adaptive scheduling and resource management are critical, can be good candidates for transferring such LLMs' capabilities. 
VAEs have proven their value in precision control and adaptability within robotics, which can be extended to manufacturing systems. 
Meo et al. demonstrate how VAE-based controllers enable precise torque control in high-precision robotic tasks \cite{meo2021multimodal}. 
This precision control approach can be transferred to manufacturing workflows requiring fine-grained anomaly detection and process parameter optimization.

These applications of generative ML in related domains provide compelling evidence for their transferability to manufacturing control.
These validated capabilities collectively establish a transferable pathway for generative ML adoption in adaptive manufacturing systems.

\subsection{Research Gaps}
Although generative ML integrations with adaptive control in manufacturing have significant potential, there are three critical gaps emerge from current approaches: 
(1) They primarily produce predictive outputs that serve as inputs to separate control systems rather than directly producing control strategies themselves. 
While limited examples of direct policy generation exist, most frameworks maintain separation between generative components and control decision-making, limiting the potential for genuinely adaptive manufacturing systems; 
(2) There is a lack of successful transferring generative approaches from related domains to manufacturing. 
Unlike fields where behavioral cloning has proven effective \cite{florence2022implicit}, manufacturing processes demand models capable of understanding underlying physical phenomena specific to manufacturing processes. 
Present approaches inadequately incorporate essential physical comprehension, prediction, and analysis of manufacturability, instead relying primarily on pattern mimicry without deeper process understanding; 
and (3) Domain adaptation challenges remain prominent, particularly due to the computational demands conflicting with real-time manufacturing requirements and methodological fundamental domain adaptation barriers.
These models were initially developed for other domains, such as image generation and language processing, causing significant difficulties in incorporating manufacturing-specific constraints, physical principles, and quality requirements necessary for robust, reliable control in production environments.

\section{FUTURE RESEARCH AND CONCLUDING REMARKS}
This review has demonstrated the significant potential of generative ML technologies in enhancing adaptive control for dynamic manufacturing processes. 
This study identified current ML-enhanced methods, generative ML's unique capabilities in uncertainty modeling, high-fidelity simulation, and sequence processing that align well with manufacturing control requirements. 
Despite promising potential in adaptive manufacturing control, three critical research gaps limit broader application: the separation between generation and control functions, insufficient physical understanding of manufacturing phenomena, and challenges adapting models designed for other domains to manufacturing-specific contexts.

Future research should focus on four strategic directions to overcome current limitations: 
(1) Developing integrated frameworks that generative models produce control policies rather than just process predictions, creating hybrid approaches that combine the predictive power of Prediction-Based methods or Quality Inference capability with the direct action capabilities of Direct Policy control, enabling responsive adaptation to manufacturing conditions; 
(2) Creating Knowledge-Integrated generative architectures that incorporate manufacturing principles and domain knowledge as explicit constraints, moving beyond pattern imitation toward process understanding; 
(3) Designing purpose-built generative models for manufacturing control applications instead of adapting architectures optimized for other domains; 
and (4) Implementing model compression and architectural improvements to reconcile computational demands with real-time processing requirements in manufacturing environments.

Integrating Generative ML into adaptive control will transform manufacturing from reactive to predictive approaches, enabling simultaneous optimization of quality, efficiency, and adaptability to dynamic conditions.



\nocite{*}

\bibliographystyle{asmeconf}  



\end{document}